\documentclass[11pt,a4paper]{article}
\usepackage{booktabs}
\usepackage{multicol,multirow}
\usepackage{graphicx}
\usepackage{enumitem}
\usepackage{numprint}
\usepackage{relsize}
\usepackage{subfig}
\usepackage[hyperref]{acl2021}
\usepackage{cleveref}
\usepackage{times}
\usepackage{latexsym}

\usepackage{comment}
\usepackage{color,colortbl}
\usepackage{microtype}

\aclfinalcopy 

\title{
    ``\textit{Definition Modeling} : \texttt{To model definitions.}'' \\
    Generating Definitions With Little to No Semantics
}

\author{Vincent Segonne\thanks{\hskip0.5em Equal contribution.} \\
  Université Grenoble Alpes \\
  \texttt{vincent.segonne}\\\texttt{@univ-grenoble-alpes.fr} \\\And
  Timothee Mickus\footnotemark[1] \\
  Helsinki University \\
  \texttt{timothee.mickus}\\\texttt{@helsinki.fi} \\}

\date{}

\begin{document}

\maketitle
\begin{abstract}
Definition Modeling, the task of generating definitions, was first proposed as a means to evaluate the semantic quality of word embeddings---a coherent lexical semantic representations of a word in context should contain all the information necessary to generate its definition.
The relative novelty of this task entails that we do not know which factors are actually relied upon by a Definition Modeling system.
In this paper, we present evidence that the task may not involve as much semantics as one might expect: we show how an earlier model from the literature is both rather insensitive to semantic aspects such as explicit polysemy, as well as reliant on formal similarities between headwords and words occurring in its glosses, casting doubt on the validity of the task as a means to evaluate embeddings.
\end{abstract}

\section{Introduction}\label{sec:intro}

Definition Modeling \citep[DefMod]{noraset-etal-2017-definition} is a recently introduced NLP task that focuses on generating a definition gloss given a term to be defined; most implementations rely on an example of usage as auxiliary input \citep[a.o.]{ni-wang-2017-learning,gadetsky-etal-2018-conditional,mickus-etal-2019-mark}.
In the last few years, it has been the focus of more than a few research works: datasets have been proposed for languages ranging from Japanese \citep{huang-etal-2022-jade} to Wolastoqey \citep{bear-cook-2021-cross}, and DefMod has even been the subject of a recent SemEval shared task \citep{mickus-etal-2022-semeval}.

Practical applications for DefMod abound, from the generation of lexicographic data for low-resource languages \citep{bear-cook-2021-cross}, to  computer-assisted language learning \citep{kong-etal-2022-multitasking}, creating learners' dictionaries \citep{jiaxin-etal-2022-compiling}, and from explaining slang \citep{ni-wang-2017-learning} to clarifying scientific terminology \citep{august-etal-2022-generating}.
Yet, it was initially conceived by \citet{noraset-etal-2017-definition} as an evaluation task for word embeddings.
If a word embedding is a coherent lexical semantic representation, then it ought to contain all the information necessary to produce a coherent gloss.
Researchers have kept this semantic aspect firmly in mind: for instance,  \citet{bevilacqua-etal-2020-generationary} argue that DefMod provides a means to dispense word-sense disambiguation (WSD) applications from fixed, rigid sense inventories.
More broadly, dictionaries in NLP are often used to capture some aspect of semantics. 

This point bears closer inquiry.
One may expect that writing definitions requires some knowledge of the meaning of the headword, but little has been done to confirm this expectation.
Here, we focus on empirically verifying what impacts a model's ability to generate valid definitions.
As such, our interest lies mostly in examining what factors in the performance of a successful Definition Modeling system, rather than in the engineering aspects of DefMod implementations.
We therefore re-purpose the fine-tuning protocol of \citet{bevilacqua-etal-2020-generationary} 
to train a BART model \citep{lewis-etal-2020-bart} to generate definitions, which we subsequently evaluate on infrequent words:
 As \citeauthor{bevilacqua-etal-2020-generationary} have extensively demonstrated the quality of their model on English data, it is suitable for our own endeavor.

Our findings suggest that it is possible to generate definition with little semantic knowledge:
Our DefMod system, far from manipulating semantic information, mostly relies on identifying morphological exponents and tying them to lexicographic patterns.
Semantic aspects of the headword---e.g., its polysemy or frequency---do not appear to weigh on model performances as captured through automatic metrics.

\section{Related Works} \label{sec:sota}

There is a broad domain of research that focuses on NLP solutions to lexicography problems and assessing how suitable they are \citep[e.g.,][]{ELX08-026,frankenberg-garcia_2020,10.1093/ijl/ecaa022,Hargraves2021-ji}. 
Conversely, many NLP works have used dictionaries to address semantic tasks, such as hypernym or synonym detection \citep{chodorow-etal-1985-extracting,gaume-etal-2004-word} word-sense-disambiguation \citep{lesk-1986,muller-etal-2006-synonym,segonne-etal-2019-using}, compositional semantics \citep{zanzotto-etal-2010-estimating,hill-etal-2016-learning-understand,mickus-etal-2020-meaning}, interpretability \citep{chang-chen-2019-word}, representation learning \citep{bosc-vincent-2018-auto,tissier-etal-2017-dict2vec} or word retrieval \citep[a.k.a. reverse dictionaries]{10.1007/978-981-15-1718-1_11}.
We more narrowly concerned ourselves with definition modeling \citep{noraset-etal-2017-definition},  formulated as a sequence-to-sequence task \citep{ni-wang-2017-learning,gadetsky-etal-2018-conditional,mickus-etal-2019-mark}. 
Our fine-tuning approach is borrowed from \citet{bevilacqua-etal-2020-generationary}; note that \citet{huang-etal-2021-definition} also employed a PLM \citep[viz. T5,][]{JMLR:v21:20-074}.
We refer readers to \citet{gardner-etal-2022-definition} for a more thorough introduction. 


\section{Model \& dataset} \label{sec:methodo}

\paragraph{Datasets} We retrieve data from DBnary \citep{serasset-2014-dbnary},\footnote{
    \url{http://kaiko.getalp.org/about-dbnary/}
} an RDF-formatted dump of Wiktionary projects.\footnote{
    \url{http://wiktionary.org/}
}
This source of data has previously been used to build DefMod datasets \citep{mickus-etal-2022-semeval}, and is available in multiple languages---a desirable trait for future replication studies.
More details are provided in \Cref{adx:preproc}.
For each term to be defined, we also tabulate its number of occurrences by tallying the number of string matches in a random subset of 5M documents from the deduplicated English Oscar corpus \citep{OrtizSuarezSagotRomary2019}.

Headword frequency is worth focusing on, for at least two reasons.
First, lexicographers are more likely to cover frequent words: dictionary-makers often espouse a data-driven approach to determine whether words should be included in general or specialized dictionaries \citep{hartmann_1992,10.1093/ijl/ecaa022};\footnote{
    Lack of corpus evidence may also be reason enough for lexicographers to ignore rarer words \citep{hanks2009impact,hanks2012}.
    Dictionaries often rely on usage data to select entries (e.g., \url{https://www.merriam-webster.com/help/faq-words-into-dictionary}) 
}
Second, dictionary users should also be less familiar with rarer words---and likely require definitions.
%
Hence, we set aside definitions where the headword has five or fewer occurrences in our Oscar subset for test purposes only, and further distinguish low-frequency headwords depending on whether they are attested in our Oscar sample.
Remaining headwords are then split 80--10--10 between train, validation, and a second held out test set, so as to also measure models on identically distributed items.
As such, we have three test sets, distinguished by the frequency of the headword in our Oscar sample: We note as  $\#=0$ the test set comprised of forms unattested in the sample; $\#\le 5$ corresponds to headwords with five or fewer occurrences; $\#>5$ matches with train set and validation set conditions.

\paragraph{Model} The core of our methodology is borrowed from \citet{bevilacqua-etal-2020-generationary}: we fine-tune a generative pretrained language model, namely BART \citep{lewis-etal-2020-bart}, to produce an output gloss given an input example of usage, where the term to be defined is highlighted by means of special tokens \texttt{<define>} and \texttt{</define>}.
We justify our adoption of their methodology by the fact that they report high results, through extensive NLG and WSD evaluation: as such, the approach they propose is representative  of successful modern approaches to DefMod, and is suitable for a study such as ours.
We refer the reader to their paper and \Cref{adx:hparams} for details.

We expect DefMod systems to be sensitive to the variety of examples of usages and number of target glosses: more examples of usage should lead to higher performances, whereas not exposing the model to polysemy should be detrimental.
This can be tested by down-sampling the training set, so as to select one gloss per headword (1G or $\forall$G) and/or one example of usage per gloss (1E or $\forall$E). 
This leads us to defining four related models: $\forall$G$\forall$E, $\forall$G1E, 1G$\forall$E, and 1G1E. 
\footnote{
    Using this notation, 1G$\forall$E means that, for a given headword, we randomly selected one gloss with all its corresponding examples; for $\forall$G1E, all glosses were considered but with only one randomly selected example for each.
}
\section{Impact of frequency, polysemy and contextual diversity}

\begin{table}[ht]
    \centering
\resizebox{\columnwidth}{!}{%
    \begin{tabular}{c r r r r}
    \toprule
    \multirow{2}{*}{\textbf{Config}} & \multicolumn{4}{c}{\textbf{Split}} \\
                                     & \textbf{Val.} & \textbf{$\#>5$} & \textbf{$\#\le 5$} & \textbf{$\#=0$} \\
    \midrule
     \textbf{$\forall$G$\forall$E} & 9.07 &  9.13 & 11.15 & 10.85 \\
     \textbf{$\forall$G1E} & 9.06 & 9.10 & 11.11 & 10.94 \\
     \textbf{1G$\forall$E} & 8.29 & 8.32 & 10.69 & 10.53 \\
     \textbf{1G1E} & 8.49 & 8.53 & 11.06 & 10.87 \\
     \bottomrule
    \end{tabular}
}
    \caption[XXX]{
        Average BLEU performances on held-out sets. Averaged on 5 runs; std. dev. $ <\pm 0.001$ always.
    }
    \label{tab:bleus}

\end{table}

Corresponding results in terms of BLEU, shown in \Cref{tab:bleus}, are in line with similar results on unseen headwords e.g. in \citet{bevilacqua-etal-2020-generationary}.\footnote{
    We observed similar patterns with most widely-used automatic NLG metrics, and focus on BLEU in the present article for brievity. 
    Nonetheless, see e.g. \citet{roy-etal-2021-reassessing} for a discussion of the limitations of this metric.
}
They also highlight a strikingly consistent behavior across all four configurations:
Mann-Whitney U tests stress that we do not observe lower performances for rarer words, as one would naively expect, except in few cases ($\forall$G$\forall$E, $\forall$G1E and 1G1E models, when comparing unattested and rare headwords) with relatively high p-values given the sample sizes ($p > 0.01$ always).

Another way to stress the lack of effect related to explicit polysemy or contextual diversity consists in correlating BLEU scores across models: Comparing the BLEU scores obtained by one model (say the $\forall$G$\forall$E) to those of another model (e.g., the 1G1E model) indicates whether they behave differently or whether BLEU scores are distributed in roughly the same fashion.
We systematically observe very high Pearson coefficients ($0.82 < r < 0.90$). 
In other words, definitions that are poorly handled in any model will in all likelihood be poorly handled in all other models, and definitions that are easy for any single model will be easy for all other models.
We provide a breakdown per split and per model in \Cref{adx:bleu-corr}, \Cref{tab:BLEU correlation}.


\section{Digging further: manual evaluation}

To better understand model behavior, we sample 50 outputs of the $\forall$G$\forall$E model, per BLEU quartile, for the validation split and our three test splits. We then annotate these 800 items as  follows. 
\subsection{Annotation scheme}

Sample items for all annotations are provided in \Cref{tab:annots-example}.

\begin{table*}[t]
    \centering
    \resizebox{\linewidth}{!}{
    \begin{tabular}{l p{6cm} p{6cm} p{4cm} r l}
    \toprule
        \textbf{POS} & \textbf{example} & \textbf{target} & \textbf{hypothesis} & \multicolumn{2}{c}{\textbf{annotation}}  \\
    \midrule \rowcolor{lightgray}
        verb & Thus was th' accomplish'd squire \textbf{endued} / With gifts and knowledge per'lous shrewd . & To invest (someone) with a given quality, property etc.; to endow. & (obsolete, transitive) To supply; to supply; to supply. & FL & 1 \\
        noun & The wealth of those societies in which the capitalist \textbf{mode of production} prevails, presents itself as “an immense accumulation of commodities,” its unit being a single commodity. & (Marxism) A combination of productive forces such as labour power and means of production, and social and technical relations of production such as property, power, laws and regulations, etc. & (economics) The economic system in which the production of goods and services is based on the production of commodities. & FL & 3 \\
        \rowcolor{lightgray}
        noun & Often, though, a \textbf{suki} to the chest will cause the sword to become lodged between bone and cartilage making it very difficult to quickly remove. & (martial arts) An opening to the enemy; a weak spot that provides an advantage for one's opponent. & (historical) A blow made by a sword to the chest. & FL & 5 \\ \midrule
        verb & 
        [...] 
        the higher of them can never \textbf{abut on} a pyknon in the case envisaged here, where the tone is introduced to disjoin the tetrachords. & (transitive) To border on. & (music, transitive) To play (a note) at the same time. & FA & 1 \\
        \rowcolor{lightgray}
        noun & “\textbf{Kurkuls} are our enemy,” he shouted, “and we must exterminate them as a social class. 
        [...]
        & (historical) A rich or supposedly rich peasant, targeted during Soviet collectivization, especially in the context of Ukraine or Ukrainians; kulak. & (rare) A kurkul. & FA & 3 \\
        adj. & And its success or failure is likely to tell whether talents [...] 
        make new fortunes from the \textbf{nonentertainment} companies that are looking to Hollywood. & Not of or pertaining to entertainment. & Not entertainment. & FA & 5 \\ \midrule
        \rowcolor{lightgray}
        adj. & an \textbf{arrant} knave, arrant nonsense & (chiefly, with a negative connotation, dated) Complete; downright; utter. & (obsolete, transitive) To make up; to invent; to invent. & PA & 0 \\
        noun & 
        [...] Another is to ban planned \textbf{obsolescence}, so manufacturers can’t create products that are designed to fail . & (uncountable) The state of being obsolete---no longer in use; gone into disuse; disused or neglected. & The state or condition of being obsolescent. & PA & 1 \\ \midrule
        \rowcolor{lightgray}
        noun & A canister of flour from the kitchen had been thrown at the looking-glass and lay like trampled snow over the remains of a decent blue suit with the \textbf{lining} ripped out which lay on top of the ruin of a plastic wardrobe. & A covering for the inside surface of something. & The outer layer of a garment. & PB & 0 \\
        adj. & an \textbf{obliquangular} triangle & (archaic, geometry) Formed of oblique angles. & (geometry) Of or pertaining to an oblique angle & PB & 1 \\ \bottomrule
    \end{tabular}
    }
    \caption{Example of annotated items. Word being defined in \textbf{bold} in the example of usage.}
    \label{tab:annots-example}
\end{table*}

\paragraph{Fluency (FL)} measures if the output is free of grammar or commonsense mistakes.
For instance, ``\texttt{(intransitive) To go too far; to go too far.}'' is rated with a FL of 1, and ``\texttt{(architecture) A belfry}'' is rated 5. 

\paragraph{Factuality (FA)} consists in ensuring that generated glosses contain only and all the facts relevant to the target senses.
Hence the output ``\texttt{Not stained.}'' generated for the headword \textit{unsatined} is annotated with a FA of 1, whereas the output ``\texttt{A small flag.}'' for the headword \textit{flaglet} is rated with a FA of 5.

\paragraph{PoS-appropriateness (PA)} A PoS-appropriate output defines its headwords using a phrase that match its part of speech---e.g., defining adjective with adjectival phrases and nouns with noun phrases.
As such, the adjective headword \textit{fried} yields the PoS-inappropriate ``\texttt{(transitive) To cook (something) in a frying pan.}'', while the production for the verb \textit{unsubstantiate}, viz. ``\texttt{(intransitive) To make unsubstantiated claims.}'' has a PA of 1.

\paragraph{Pattern-based construction (PB)} An output is said to display a pattern-based construction whenever it contains only words that are semantically tenuous or morphologically related to the headword.
The headword \textit{clacky} thus yield the PB output ``\texttt{Resembling or characteristic of clacking.}'', and the headword \textit{fare} yields the non-PB production ``\texttt{(intransitive) To do well or poorly.}''
\footnote{
    FA and FL are on a 5-point scale, PA and PB 
    are binary.
}


\subsection{Results of the manual evaluation}

When looking at all 800 annotations, we find that outputs tend to be fluent (average FL of 4.37) and overwhelmingly PoS-appropriate ( 95\%).
They frequently involve patterns (36.5\% of PBs) which often involve a straight copy of the headword (10\% of all productions).
On the other hand, factuality is lacking (average FL of 2.69).

\begin{table}[!t]
    \centering
    \npdecimalsign{.}
    \nprounddigits{3}
    \begin{tabular}{l n{1}{3} n{1}{3} n{1}{3}}
    \toprule
        \textbf{Trait} & \textbf{Cohen $\kappa$} & \textbf{Spearman $\rho$} & \textbf{Pearson $r$} \\
    \midrule
\textbf{FL} & 0.4051553205551883 & 0.6332368582148349 & 0.6929599343280817 \\
\textbf{FA} & 0.3736951983298539 & 0.7408550798026733 & 0.7677954316735561 \\ 
\textbf{PA} & 1.0 & 1.0 & 1.0 \\
\textbf{PB} & 0.7797577335068575 & 0.7836947310718164 & 0.7836947310718164 \\ 
    \bottomrule
    \end{tabular}
    \caption{Manual annotations, inter-annotator agreement. Pearson $r$ were computed on $z$-normalized annotations.}
    \label{tab:annot}
\end{table}
\paragraph{Inter-annotator agreement} To quantify how consensual our annotations are, we randomly sample 200 items for dual annotation. 
Results in \Cref{tab:annot} highlight that, while the two annotators have different sensibilities as to the magnitude of the mistakes in FL and FA (as shown by the low $\kappa$), relative judgments on fluency and factuality are consistent (as shown by $\rho$ and $r$).
Hence, we $z$-normalize FA and FL in the rest of this analysis.

\paragraph{Effects of patterns} Mann-Whitney U-tests on FA and FL annotations show that non-pattern-based outputs are statistically rated with lower FL ($p < 3 \cdot 10^{-6}$, common language effect size $f = 42.3\%$) and lower  FA ($p < 2 \cdot 10^{-9}$, $f = 37.7\%$) than pattern-based definitions, despite no significant difference in BLEU scores ($p = 0.262$).
On the other hand, BLEU scores are correlated with FL and FA ratings (Spearman $\rho = 0.094$ and $\rho = 0.276$ respectively). 
In sum, the morphologically complex nature of a headword drives much of the behavior of our DefMod system.
While BLEU captures some crucial aspects we expect to be assessed in DefMod, it is still impervious to this key factor.

\begin{table}[!t]
    \centering
    \npdecimalsign{.}
    \nprounddigits{2}
    \begin{tabular}{n{1}{2} n{1}{2} n{1}{2} n{1}{2}}
    \toprule
        \multicolumn{4}{c}{\textbf{Split}} \\
        \textbf{Val.} & \textbf{$\#>5$} & \textbf{$\#\le 5$} & \textbf{$\#=0$} \\
    \midrule
        5.602605158994286 & 5.723377820536981 & 5.110977139866202 & 4.8521161386247815 \\
    \bottomrule
    \end{tabular}
    \caption{Performances with headword ablation}
    \label{tab:masked-headword}
\end{table}

To further confirm that patterns are indeed crucial to a DefMod system's performance, we train a model on data where headwords have been removed from examples of usages, keeping the surrounding control tokens. 
This in effect creates a 2-token sentinel for which the decoder must generate a gloss, and deprives the model of information about headword form.
BLEU scores drastically drop with this ablated train set, as shown in \Cref{tab:masked-headword}.
We also find unattested headwords yielding statistically lower BLEUs than rare headwords, which in turn yield lower BLEUs than the other two splits (Mann-Whitney U tests, $p < 10^{-7}$).

\paragraph{Frequency and polysemy} We now return to polysemy and word frequency. 
We consider as an indicator of word polysemy the number of definitions for that headword present in our corpus, whereas we rely on our Oscar sample to derive frequency counts.
Frequency and definition counts appear to be highly correlated (Spearman $\rho =  0.406$), and both also anti-correlate with PB ($\rho=-0.1143$ and $\rho=-0.111$ respectively), i.e., rare, monosemous words are defined by the model with patterns (that is, they are likely morphologically complex).
We also observe an anticorrelation between FL and definition count (Spearman $\rho=-0.105$), which could be explained by the fact that patterns tend to yield more fluent outputs, as we just saw---however, as we do not observe a correlation between frequency and FL, the interaction between FL and polysemy (as measured by definition count) is likely not so straightforward.\footnote{
    Neither do we observe no correlation with FA and PA. 
}
Finally, BLEU scores do not correlate with word frequency nor definition counts, which strengthens our claim that this DefMod system makes limited use semantic information to generate glosses---if at all.

\begin{table}[ht]
    \centering
\resizebox{\columnwidth}{!}{
    \npdecimalsign{.}
    \nprounddigits{2}
    \begin{tabular}{l@{{}}n{2}{2}@{{~~}}n{2}{2}}
    \toprule
        & {{\textbf{FL}}} & {{\textbf{FA}}} \\
    \midrule
        \textbf{BertScore} \citep{Zhang2020BERTScore:} & 0.163183 & 0.371737  \\ 
        \textbf{BLEU} \citep{papineni-etal-2002-bleu} & 0.094475  & 0.276090 \\
        \textbf{chrF} \citep{popovic-2015-chrf} & {{\quad--}} & 0.350936  \\
        \textbf{GLEU} \citep{wu2016googles} & {{\quad--}} & 0.292390 \\
        \textbf{METEOR} \citep{banerjee-lavie-2005-meteor} & {{\quad--}}  & 0.314580\\
        \textbf{ROUGE-L} \citep{lin-2004-rouge} & {{\quad--}}  & 0.372419 \\
        \textbf{TER} \citep{snover-etal-2006-study} & -0.102907  & -0.267827 \\
    \bottomrule
    \end{tabular}
}
    \caption{Correlation of FA and FL with NLG metrics. Missing values correspond to insignificant coefficients. }
    \label{tab:nlg-metrics}
\end{table}

\paragraph{Alternatives to BLEU} These annotations leave one question unanswered: is BLEU an adequate means of measuring DefMod productions?
In \Cref{tab:nlg-metrics}, we compare the Spearman correlation coefficient of various NLG metrics with our FA and FL annotations.
Most NLG metrics do not correlate with fluency ratings: we posit this is due to the overwhelming majority of highly fluent productions in our sample.
As for BLEU, it doesn't produce the highest (anti-)correlations---they are instead attested with BertScore for FL and ROUGE-L for FA. 
Lastly, Mann-Whitney U tests comparing metrics with respect to PB annotations indicate that most of these are not sensitive to the presence or absence of a pattern, with the exception of chrF ($f=0.43$) and TER ($f=0.42$).
In all, our annotated sample suggests that most NLG metrics appear to display a behavior similar to BLEU: they capture factuality to some extent---but not the importance of patterns.



\section{Conclusions}

In this work, we have presented how an earlier Definition Modeling system was able to achieve reasonable performances and produce fluent outputs, although the factual validity leave much to be desired.
This behavior is almost entirely due to morphologically complex headwords, for which the model is often able to derive reasonable glosses by decomposing the headword into a base and an exponent, and mapping the exponent to one of a limited set of lexicographic patterns. 
The model we studied seems more sensitive to formal traits than to explicit accounts of polysemy.
There are numerous limitations to this work: we focused on one specific fine-tuning approach for one specific English PLM.
Nonetheless, we have shown that models can achieve reasonable performances on DefMod without relying on semantics, casting doubt on the task's usefulness for word embedding evaluation, as initially suggested by \citet{noraset-etal-2017-definition}

In other words: using lexicographic data as inputs for an NLP model does not ensure that it will pick up on the semantic aspects contained therein.

\section*{Acknowledgments}

 We thank IWCS and $\ast$SEM reviewers for their insightful comments on this work.

\vspace{0.5ex}

\noindent
{ 
\begin{minipage}{0.1\linewidth}
    \vspace{-10pt}
    \raisebox{-0.2\height}{\includegraphics[trim =32mm 55mm 30mm 5mm, clip, scale=0.18]{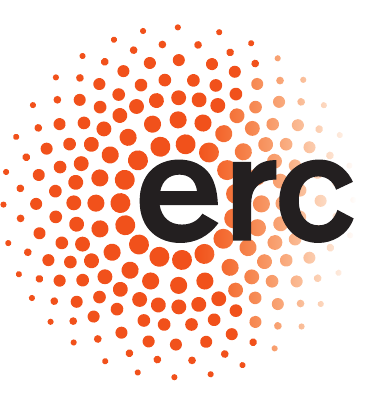}} \\[0.25cm]
    \raisebox{-0.25\height}{\includegraphics[trim =0mm 5mm 5mm 2mm,clip,scale=0.075]{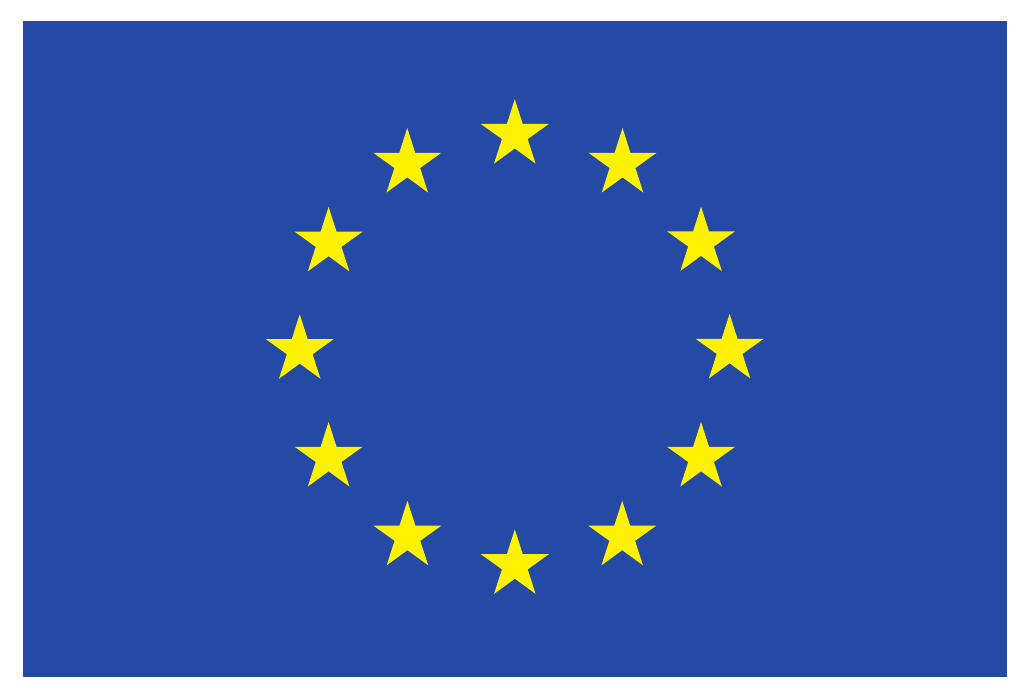}}
\end{minipage}
\hspace{0.01\linewidth}
\begin{minipage}{0.85\linewidth}
This work is part of the FoTran project, funded by the European Research Council (ERC) under the EU's Horizon 2020 research and innovation program (agreement \textnumero{}~771113). ~~We ~also ~thank ~the ~CSC-IT \vspace{0.5ex}
\end{minipage}
\begin{minipage}{\linewidth}
\noindent Center for Science Ltd., for computational resources. 
\end{minipage}%
}

\bibliographystyle{acl_natbib}
\bibliography{anthology,acl2021}

\newpage
\clearpage
\appendix

\section{Hyperparameters} \label{adx:hparams}

Models are implemented in fairseq \citep{ott-etal-2019-fairseq}. We used the \texttt{bart\_large} model and followed the instructions on the github repository for finetuning BART on the summary task.\footnote{
    \url{https://github.com/facebookresearch/fairseq/blob/main/examples/bart/README.summarization.md}
} We used the same parameters except for the learning rate, which after some experiments, was set to $5 \cdot 10^{-6}$. For every configuration ($\forall$G$\forall$E, $\forall$G$1$E,$1$G$\forall$E, $1$G$1$E) we kept the model with the best loss on the validation dataset.

\section{Data preprocessing} \label{adx:preproc}

In the present work, we retrieve definition glosses (i) associated with an example of usage and (ii) where the term to be defined is tagged as a noun, adjective, verb, adverb or proper noun.
Like \citeauthor{bevilacqua-etal-2020-generationary}, we also consider MWEs as potential terms to define.

To highlight a headword within an example of usage, the approach of \citet{bevilacqua-etal-2020-generationary} consists in surrounding them with learned task-specific control tokens.
We therefore parse example of usages using SpaCy\footnote{
    \url{https://spacy.io/}
} to retrieve the first sequence of tokens whose lemmas match with the lemmas of the term to be defined.

The BART model we fine-tune on DefMod has been pretrained on OpenWebText, which contains some pages retrieved from Wiktionary. 
We preemptively remove these pages from all dataset splits, so as to ensure there is no overlap between pre-train, train and test data.

Frequencies are tabulated on a case-folded, whitespace-normalized subset of the Oscar corpus. 
In practice, we extract the number of hard string matches of each headword preprended and appended with word boundaries.

\section{BLEU scores correlations} \label{adx:bleu-corr}

\begin{table}[!h]
    \centering
\subfloat[Validation split]{
    \npdecimalsign{.}
    \nprounddigits{2}
    \begin{tabular}{l n{1}{2}@{{~}}n{1}{2}@{{~}}n{1}{2}@{{~}}n{1}{2}}
    \toprule
                                      & \textbf{\smaller $\forall$G1E} & \textbf{\smaller 1G$\forall$E} & \textbf{\smaller 1G1E} \\
        \textbf{\smaller $\forall$G$\forall$E} & 0.8937069938524711 & 0.8670630114034488 & 0.8537759172728328 \\
        \textbf{\smaller $\forall$G1E}         &                    & 0.8446929860154532 & 0.8656924894015559  \\
        \textbf{\smaller 1G$\forall$E}         &                    &                    & 0.8760275967986277\\
    \bottomrule
    \end{tabular}
}

\subfloat[Test $\#>5$ split]{
    \npdecimalsign{.}
    \nprounddigits{2}
    \begin{tabular}{l n{1}{2}@{{~}}n{1}{2}@{{~}}n{1}{2}@{{~}}n{1}{2}}
    \toprule
                                      & \textbf{\smaller $\forall$G1E} & \textbf{\smaller 1G$\forall$E} & \textbf{\smaller 1G1E} \\
        \textbf{\smaller $\forall$G$\forall$E} & 0.8862593241734686 & 0.8550144468728926 & 0.8534479460513822 \\
        \textbf{\smaller $\forall$G1E}         &                    & 0.8258692101441409 & 0.870000585537346 \\
        \textbf{\smaller 1G$\forall$E}         &                    &                    & 0.8659820931456106 \\
    \bottomrule
    \end{tabular}
}

\subfloat[\label{tab:bleu-correl-example} Test $\#\le 5$ split]{
    \npdecimalsign{.}
    \nprounddigits{2}
    \begin{tabular}{l n{1}{2}@{{~}}n{1}{2}@{{~}}n{1}{2}@{{~}}n{1}{2}}
    \toprule
                                      & \textbf{\smaller $\forall$G1E} & \textbf{\smaller 1G$\forall$E} & \textbf{\smaller 1G1E} \\
        \textbf{\smaller $\forall$G$\forall$E} & 0.8778894896642774 & 0.8872033392835064 & 0.8581742302742222 \\
        \textbf{\smaller $\forall$G1E}         &                    & 0.8518918107435429 & 0.8840534564290756 \\
        \textbf{\smaller 1G$\forall$E}         &                    &                    & 0.8804211748104476 \\
    \bottomrule
    \end{tabular}
}

\subfloat[Test $\#=0$ split]{
    \npdecimalsign{.}
    \nprounddigits{2}
    \begin{tabular}{l n{1}{2}@{{~}}n{1}{2}@{{~}}n{1}{2}@{{~}}n{1}{2}}
    \toprule
                                      & \textbf{\smaller $\forall$G1E} & \textbf{\smaller 1G$\forall$E} & \textbf{\smaller 1G1E} \\
        \textbf{\smaller $\forall$G$\forall$E} & 0.8769118318106075 & 0.8793409354671265 & 0.854506173689362 \\
        \textbf{\smaller $\forall$G1E}         &                    & 0.8550610614504774 & 0.8816787132020953 \\
        \textbf{\smaller 1G$\forall$E}         &                    &                    & 0.8806595945140945 \\
    \bottomrule
    \end{tabular}
}
    \caption{BLEU scores correlations (Pearson $r$)}
    \label{tab:BLEU correlation}
\end{table}

In \Cref{tab:BLEU correlation}, we display how similar are the behaviors on different models across splits.
Each sub-table corresponds to a different split, and pits all combinations of models.
For instance, the last cell in the second row of sub-Table \ref{tab:bleu-correl-example} indicates that to the Pearson correlation between the $\forall$G1E and the 1G1E on the $\#\le 5$ test split is above 88.4\%.
The crucial fact that emerges from these tables is the distribution of BLEU is very similar across all models we tested---which entails that explicit polysemy or contextual diversity do not weight on performances, as measured through BLEU scores.

\end{document}